\documentclass{Interspeech2024}
\usepackage{tabularx}
\usepackage{amssymb}
\usepackage{inconsolata}

\usepackage{mathrsfs}
\usepackage{paralist}
\usepackage{booktabs}
\usepackage{multirow}
\usepackage{pgfplots}
\usepackage{scalefnt}
\usepackage{amsthm,amsmath,amssymb}
\usepackage{mathtools}
\usepackage{bm}
\usepackage{colortbl}
\usepackage{xcolor}
\usepackage{array}
\usepackage{stfloats}
\usepackage{tikz-dependency}
\usepackage{tikz}
\usepackage{siunitx}
\usepackage{graphicx}
\usepackage[ruled]{algorithm2e}
\usepackage{lineno}
\usepackage{subfigure}
\usepackage{helvet}
\usepackage{courier}
\usepackage{CJKutf8}
\usepackage{fancyvrb}
\interspeechcameraready 

\title{LLM-Driven Multimodal Opinion Expression Identification}

\name[affiliation={1,*}]{Bonian}{Jia}
\name[affiliation={1,*}]{Huiyao}{Chen}
\name[affiliation={}]{Yueheng}{Sun$^{1,\dagger}$}
\name[affiliation={2}]{Meishan}{Zhang}
\name[affiliation={2}]{Min}{Zhang}

\address{
  $^1$Tianjin University, China\\
  $^2$Institute of Computing and Intelligence, Harbin Institute of Technology (Shenzhen), China}
\email{bonianjia@tju.edu.cn, chenhy1018@gmail.com, 	yhs@tju.edu.cn,\\mason.zms@gmail.com, zhangmin2021@hit.edu.cn}


\begin{document}

\maketitle

\renewcommand{\thefootnote}{\fnsymbol{footnote}}
\footnotetext{$^*$~~Authors contributed equally.}
\footnotetext{$^\dagger$~~Corresponding author.}
\footnotetext{\texttt{https://github.com/3cqscbr26/LLM-MOEI}}
\setcounter{footnote}{0}
\renewcommand{\thefootnote}{\arabic{footnote}}

\begin{abstract}
    
Opinion Expression Identification (OEI) is essential in NLP for applications ranging from voice assistants to depression diagnosis. This study extends OEI to encompass multimodal inputs, underlining the significance of auditory cues in delivering emotional subtleties beyond the capabilities of text. We introduce a novel multimodal OEI (MOEI) task, integrating text and speech to mirror real-world scenarios. Utilizing CMU MOSEI and IEMOCAP datasets, we construct the CI-MOEI dataset. Additionally, Text-to-Speech (TTS) technology is applied to the MPQA dataset to obtain the CIM-OEI dataset. We design a template for the OEI task to take full advantage of the generative power of large language models (LLMs). Advancing further, we propose an LLM-driven method STOEI, which combines speech and text modal to identify opinion expressions. Our experiments demonstrate that MOEI significantly improves the performance while our method outperforms existing methods by 9.20\% and obtains SOTA results.

\end{abstract}

\section{Introduction}

Opinion expression identification (OEI) is a critical task of opinion mining in natural language processing (NLP), initially introduced by Breck et al. \cite{DBLP:conf/ijcai/BreckCC07}.
It aims to concurrently recognize the text spans that express particular opinions and their sentiment polarity, facilitating a lot of sentiment-related downstream applications such as voice assistants \cite{ma2023emotion}, public sentiment monitoring \cite{DBLP:conf/acl/ZhangXSZWZ22, DBLP:journals/corr/abs-2305-14842} and depression diagnosis \cite{10.1145/2661806.2661816}.
Neural network-based models for OEI have emerged as the predominant approach.
Irsoy and Cardie \cite{DBLP:conf/emnlp/IrsoyC14} significantly improved the extraction of opinion viewpoints by employing deep bidirectional recurrent neural networks (RNNs), thereby setting new performance standards.
Zhang et al. \cite{DBLP:journals/is/ZhangWF19} introduced the use of bidirectional long short-term memory (BiLSTM) networks for sentence encoding and incremental decoding processes.

\begin{figure}[t]
    \centering
    \includegraphics[width=0.9\columnwidth]{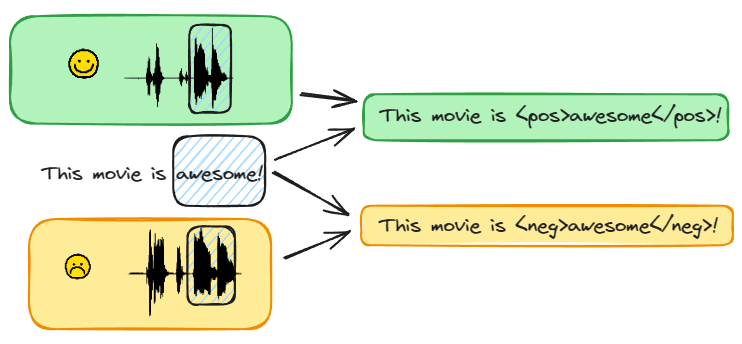}
    \caption{Opinion expressions in the same sentence show different emotional polarity in different speech scenarios.}
    \label{fig:intro}
\end{figure}

Following the introduction of BERT \cite{DBLP:conf/naacl/DevlinCLT19}, Xia et al. \cite{DBLP:conf/naacl/XiaZWLZHSZ21} proposed a novel span-based end-to-end approach, employing BERT for encoding and the BiLSTM model to improve performance.
Subsequently, the field of OEI has increasingly been dominated by pre-trained models.
Recently, with the emergence of large language models (LLMs) exhibiting superior capabilities, Wu et al. \cite{wu2023event} innovated by reformulating the extraction of opinion targets into a question answering (QA) framework.
This approach involves the use of LLMs to manage opinion mining tasks through the creation of context-specific prompts.
Nonetheless, the application of LLMs in OEI remains unexplored.
On the other hand, integrating multimodal features, particularly auditory cues, has shown to be advantageous for OEI.
Speech conveys more nuanced emotional content than text \cite{wang2023leveraging}, as the emotional tone can significantly alter the sentiment polarity of the same text, as illustrated in Figure \ref{fig:intro}.
Moreover, inherent speech features like emphasis and pauses are critical in enhancing OEI task accuracy.
These auditory characteristics provide supplementary insights absent in text, thereby facilitating a more precise interpretation of the opinion expressions.

In this work, we introduce a multimodal OEI (MOEI) task, focusing on the integration of text and speech to construct authentic OEI scenarios.
Acknowledging the inherent challenge presented by real-world speech, which frequently contains interjections and noise, thereby complicating perfect alignment with text, we utilize the open-source, multimodal film and video datasets CMU MOSEI \cite{zadeh2018multimodal} and IEMOCAP \cite{busso2008iemocap} to create the CI-MOEI dataset featuring this imperfect alignment.
To further enhance our dataset, we apply text-to-speech (TTS) technology to generate synthesized speech for the extensively utilized OEI dataset, MPQA. This synthesized speech integrates as an augmented component of our dataset, culminating in the creation of the CIM-MOEI dataset.
It aims to test whether MOEI data from laboratory environments can serve as effective training material to enhance performance.

\begin{figure*}[t]
  \centering
  \includegraphics[width=0.8\textwidth]{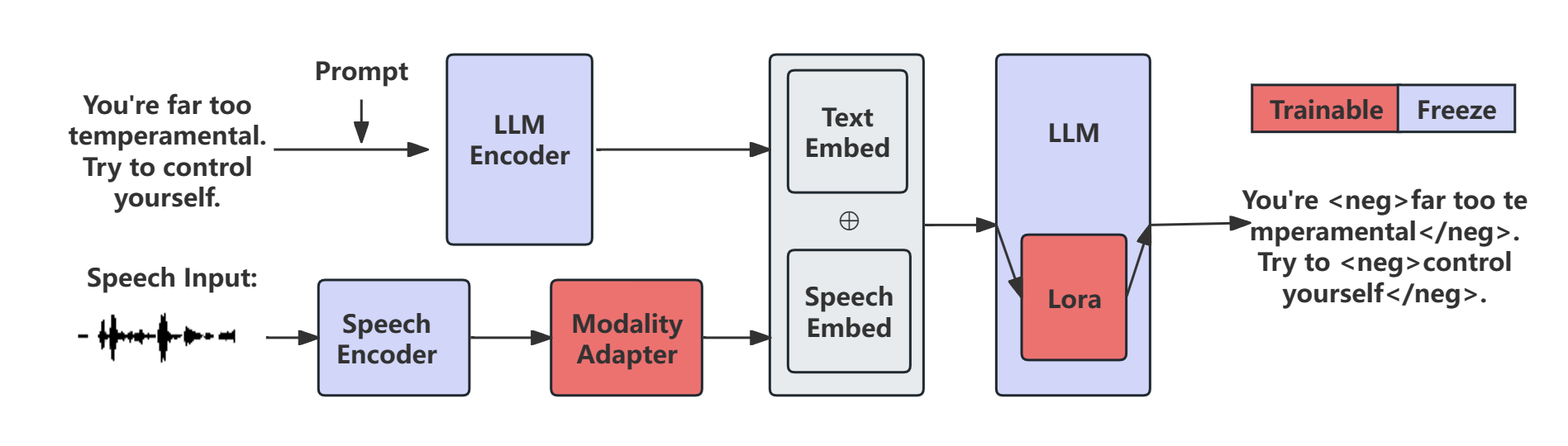}
  \caption{The overall architecture of our method STOEI, where $\oplus$ indicates vectorial concatenation.}
  \label{fig:pipeline_draw}
\end{figure*}

LLMs have demonstrated significant potential in sentiments analysis \cite{zhang2023sentiment,sun2023sentiment}, motivating us to expand the LLMs-based approach for OEI.
First, we design output templates uniquely suited to the OEI task, leveraging the advanced processing power of LLMs.
These templates are engineered to facilitate the simultaneous output of all prediction results, thereby enhancing the efficiency.
To further refine our approach, we introduce a novel LLM-driven multimodal methodology,
designated as STOEI,
which combines speech and text modal information to identify opinion expressions.
We encode speech into speech features, which are then mapped to the textual vector space through an Adapter.
By concatenating these speech features with textual vectors, we enable LLMs to effectively process both speech and textual information.


We conduct experiments on these two datasets, and the results demonstrate a significant enhancement in the performance of MOEI when utilizing both text and speech inputs, compared to traditional unimodal (text-only) inputs. Furthermore, our proposed method surpasses existing multimodal techniques by 9.20\%, achieving state-of-the-art (SOTA) results.
Our code and datasets will be publicly available to facilitate future research.

\section{Dataset}

In this paper, we propose a novel task named MOEI, focusing here on the case involving two modalities, speech and text, with the aim of combining speech and text information to further improve the performance of the OEI task.
Data selection as well as dataset construction is performed to fulfill the task requirements. Next, we will describe these two steps in detail.

\subsection{Data Selection}

In multimodal research, the conventional assumption of perfect speech-text alignment does not reflect real-world complexities, where issues like vocal interference, audio offset and transcription errors are common.
To address this, our study explores the inherent misalignments by utilizing multimodal emotion datasets like CMU MOSEI, featuring manually transcribed YouTube clips, and IEMOCAP, with actor performances that diverge from scripts.
These choices highlight the challenges of accurate speech-text synchronization and introduce realistic noise into our analysis.
The audio of the above two datasets are derived from real scenes.
In addition, we based our speech synthesis on the MPQA dataset, which is commonly used in OEI tasks, and our MOEI annotation guide is inspired by MPQA.

\subsection{Dataset Construction}

For multimodal emotion datasets, we first establish annotation guidelines aligned with those of the MPQA, aimed at identifying sentiment expressions and their corresponding polarities within sentences.
Then, we recruit three university students with majors in English and one native English speaker to serve as annotators.
For each piece of data, the three student annotators provide their assessments, and the native English speaker makes the final decision on which annotation to adopt.
This process results in an agreement rate of 81\% among the student annotators.
In instances of uncertainty, the disputed item is referred to an expert in OEI for final determination.

For the OEI dataset MPQA, we select sentences that are labeled with sentiment polarity tags 'POS' or 'NEG' and remove any unspecified content.
Subsequently, we employ TTS functionality of Azure \footnote{\href{https://learn.microsoft.com/en-us/azure/ai-services/speech-service/text-to-speech}{https://learn.microsoft.com/en-us/azure/ai-services/speech-service/text-to-speech}} as the tool for speech synthesis.
To diversify the synthesized voice, five different speakers and ten types of vocal emotions are chosen. These selections are allocated according to the sentiment polarity of the sentences.

\section{Method}


\subsection{LLM-driven OEI} 

Traditionally, OEI tasks have typically used a sequence labeling approach. Recently, some researchers have begun to explore the use of LLM for emotion recognition in conversation \cite{DBLP:journals/corr/abs-2309-11911}.
Nevertheless, this method does not fully exploit the sequence generation potential of LLMs. We advocate for a novel LLM-driven OEI template, which enables the extraction of all opinion expressions along with their respective sentiment polarities.

And then, we redesign the output template of the OEI task. In the original sentence, we add $\langle tag \rangle \langle /tag \rangle$ to mark the emotional opinion words as shown in the figure.
Among them, $\langle tag \rangle$ represents the label start, $\langle /tag \rangle$ represents the label end, and the value of the tag is 'pos' or 'neg', which is used to represent the polarity of emotion.
An example of the final output is shown on the right side of Figure \ref{fig:pipeline_draw}.

\subsection{STOEI} 

It is difficult to identify expressions containing complex emotions based on text alone. In addition, a sentence can also express multiple emotions, and different emotional expressions have different meanings. Speech representations can often convey clues about emotional information. Therefore, we introduce STOEI, a novel LLM-driven method for the MOEI task that integrates four key components: an LLM encoder, a speech encoder, a modality adapter, and an LLM. The comprehensive architecture of our approach is depicted in Figure \ref{fig:pipeline_draw}.



Formally, we have a piece of raw speech $U$, and its corresponding text $S$.
We add a prompt before the text $S$ for task definition.
Thus, we obtain a new text $X=(w_1, w_2, \ldots, w_n)$, where $w$ denotes a word.
Next, we can get the text embedding representation $\bm{H}^{\text{text}}$ of $X$ by using an LLM encoder.
Similarly, we input the piece of raw speech $U$ into the speech encoder and extract speech feature embedding $\bm{A}$ from the final hidden layer.

The core STOEI lies in jointly exploiting text and speech modalities, underpinned by the modality adapter. This integration leverages an architecture inspired by the BLSP framework \cite{wang2023blsp}, comprising three one-dimensional convolutional layers \cite{hannun2019sequence} followed by a bottleneck layer \cite{DBLP:conf/icml/HoulsbyGJMLGAG19}. These convolutional layers serve the primary purpose of sub-sampling, which is critical for distilling speech features to an optimal context size conducive to analysis. Following this, the bottleneck layer introduces additional adaptive layers, significantly boosting the model’s ability to both assimilate and interpret nuances from speech data.
We use the modality adapter to map speech features to the same vector space as text features, and we get the speech embedding representation $\bm{H}^{\text{speech}}$.
The whole process can be formalized as follows:
\begin{equation}
    \bm{H}^{\text{speech}} = \text{Adapter}(\bm{A}).
\end{equation}



Then, we perform a vectorial concatenation operation on the text embedding $\bm{H}^{\text{text}}$ and the speech embedding $\bm{H}^{\text{speech}}$ to obtain a combined representation $\bm{H}$ that contains multimodal information, which can be represented as:
\begin{equation}
        \bm{H} = \bm{H}^{\text{text}} \oplus \bm{H}^{\text{speech}},
\end{equation}
where $\oplus$ indicates vectorial concatenation.

After that, we input the combined representation $\bm{H}$ into the LLM to obtain the output probability distribution $\bm{O}$.
For training, we use the lora module to decompose the LLM parameter matrix into two low-rank matrices multiplied together, reducing the number of parameters needed for fine-tuning.
We exploy cross-entropy loss as our training objective:
\begin{equation}
\mathcal{L} = -\frac{1}{N} \sum_{i=1}^{N} \sum_{t=1}^{T} \sum_{c=1}^{M} \bm{y}_{i,t,c} \log(\text{softmax}(\bm{O})_{i,t,c}),
\end{equation}
where $\bm{y}_{i,t,c}$ is an one-hot matrix, for sequence $i$ at time step $t$, it is 1 if the true next word is $c$, 0 otherwise.

\section{Experiments}

In this section, we present the dataset construction process, the results compared with other approaches and the related analysis.


\subsection{Dataset details}

Table \ref{tab:data} reports the statistics of datasets.
The CI-MOEI and Test are authentic speech-based MOEI datasets developed through the annotation of the \textbf{C}MU MOSEI and \textbf{I}EMOCAP.
The CIM-MOEI refers to an extension of the CI-MOEI, additionally incorporating synthesized speech generated from the MPQA using TTS technology. The CI-OEI and CIM-OEI have been derived by removing the speech components from their respective datasets, retaining only the textual modality data.

\begin{table}[t]

  \centering
\caption{Statistics of Datasets. \#Sent, \#POS, and \#NEG denote the number of sentences, positive opinion expressions, and negative opinion expressions in the dataset, respectively.}
\label{tab:data}
\small
  \setlength{\tabcolsep}{6pt}
\renewcommand{\arraystretch}{1.5} 
\resizebox{0.8\columnwidth}{!}{
  \begin{tabular}{c c c c c}
    \toprule
    \textbf{Dataset}& \textbf{\#Sent}& \textbf{\#POS}& \textbf{\#NEG}& \textbf{Minutes}\\
    \hline
    CI-MOEI& 1739& 1226& 1591&187.7\\

    CIM-MOEI& 8592& 5241& 8800& 1273.0\\
    \hline
    Test& 773& 454& 480&468.4\\
    \bottomrule
  \end{tabular}
}

\end{table}

\subsection{Experiments Setup}



\textbf{Model details.}
Regarding speech encoder, the Whisper-large-v2 \cite{radford2022whisper} is selected due to its commendable performance in speech recognition tasks and its capability in emotion recognition.
Regarding the large language model, the Vicuna 7B v1.5 \cite{vicuna2023} is chosen as the experimental base model, and we use the encoder of Vicuna as the text encoder.
Vicuna, a conversational assistant, is fine-tuned on shared user dialogues collected from ShareGPT using LLaMA2. The selection of Vicuna is motivated by its enhanced emotional intelligence \cite{wang2023emotional}. 

\textbf{Baseline.}
In the context of speech perception comparative analysis, the LLaMA2 7B \cite{touvron2023llama} is also selected for evaluation.
The text-based baseline models subject to comparison include GPT-4 and Bert-BiLSTM-CRF \cite{zhang-etal-2022-identifying}.
GPT-4 is the best text-generation LLM.
Bert-BiLSTM-CRF is a representative approach for traditional classification models.
We also compare an LLM using cross-modal fusion technology, Whispering-LLaMA \cite{radhakrishnan-etal-2023-whispering}.
It fuses the audio features and the Whisper linear layer into the LLaMA model, injecting it with audio information, while the Whisper linear layer generates key-value pairs in the cross-attention mechanism of the decoder.
To better adapt to the needs of MOEI, we replace the base model of Whispering-LLaMA with Vicuna.

\textbf{Training details.} The entirety of our experiments is conducted utilizing an assembly of eight NVIDIA A100 Tensor Core GPUs, with a capacity of 40G. The models is fine-tuned over 15 epochs, employing a learning rate of 5e-5, with a batch size set to 8. The epoch size is determined based on optimal training outcomes. For optimization, the AdamW optimizer is the selected mechanism. About the Low-Rank Adaptation module, the rank of adaptation is established at 8, with the scaling factor for the adaptation set at 16, and a dropout rate of 0.05.
By default, the weights of Whisper and LLaMA2 are frozen. The only parts available for training are the modality adapter and the LoRA module.

\subsection{Evaluation Metrics}


In OEI, model accuracy is determined by an exact match of \((\text{start position}, \text{end position}, \text{sentiment polarity})\) tuples against gold-standard labels, requiring identical positions and sentiment polarity. 
Precision (P), recall (R), and the F$_1$ score are computed to provide a comprehensive evaluation of the model.

\subsection{Main Results}
\begin{table}[ht]
  \caption{The experiment results of traditional OEI methods and speech perception method. The units in the table are \%.}
  \label{tab:ssom_performance}
  \centering
  \setlength{\tabcolsep}{5pt}
\renewcommand{\arraystretch}{1.5} 
\resizebox{0.8\columnwidth}{!}{
  \begin{tabular}{c c c c c}
    \toprule
    \textbf{Method}& \textbf{Dataset} & \textbf{P}& \textbf{R}& \textbf{F1}\\
    \hline
 \multicolumn{5}{c}{\textbf{Traditional Text-based OEI}}\\
    \midrule
    Bert-BiLSTM-CRF& CI-OEI&  49.24& 49.45& 49.34\\
    GPT4              & --- & 79.74& 64.95& 71.63\\
    Vicuna            & CI-OEI& \textbf{72.60}& \textbf{73.85}& \textbf{73.22}\\
    \hline
 \multicolumn{5}{c}{\textbf{Text-Speech based MOEI}}\\
 \hline
 Whispering-LLaMA & CI-MOEI& 76.14& 74.92&75.53\\
    STOEI-Vicuna& CI-MOEI& \textbf{84.14}& \textbf{85.32}& \textbf{84.73}\\
    \bottomrule
  \end{tabular}
  }
\end{table}

\textbf{Speech perception.} Analysis of Table \ref{tab:ssom_performance} shows that the F1 score of GPT4 increases by 22.29\% compared to Bert-BiLSTM-CRF when only text input is available, indicating the potential of LLM in OEI. Meanwhile, we find that Vicuna fine-tuned with CI-OEI can exceed the F1 score of GPT4 by 1.59\%, which proves that LLM fine-tuning is a promising solution.

Upon combining the auditory modality, both Whispering-LLaMa and STOEI outperform Vicuna, demonstrating that perceiving non-aligned speech indeed aids LLMs in OEI. STOEI surpasses Whispering-LLaMA by a 9.20\% margin in terms of the F1 score. These findings highlight the superiority of the modality combination approach employed by our proposed STOEI, underscoring its effectiveness in aiding OEI through the combination of speech modalities.

\begin{table}[ht]

  \centering
\caption{The experimental results on the CIM-MOEI dataset. $\dagger$ indicates that the base model has been fine-tuned on CIM-OEI.}
\label{tab:enhanced_ssom_performance}
\small
  \setlength{\tabcolsep}{5pt}
\renewcommand{\arraystretch}{1.5} 
\resizebox{0.8\columnwidth}{!}{
  \begin{tabular}{c c c c c}
    \toprule
    \textbf{Method}& \textbf{Dataset} & \textbf{P} & \textbf{R} & \textbf{F1} \\
    \hline
 \multicolumn{5}{c}{\textbf{Traditional Text-based OEI}}\\
 \hline
 Bert-BiLSTM-CRF& CIM-OEI& 51.09& 54.68&52.83\\
    Vicuna              & CIM-OEI& 76.52& 72.67& 74.55\\
    \hline
 \multicolumn{5}{c}{\textbf{Text-Speech based MOEI}}\\
 \hline
 Whispering-LLaMA& CIM-MOEI& 76.96& 76.63&76.8\\
    STOEI-Vicuna$\dagger$& CIM-MOEI& 86.45& 86.17& 86.36\\
    STOEI-Vicuna& CIM-MOEI& \textbf{90.50}& \textbf{89.60}& \textbf{90.09}\\
    \bottomrule
  \end{tabular}
  
}

\end{table}

\textbf{Further research the impact of speech and text.}
From Table~\ref{tab:enhanced_ssom_performance}, it becomes clear that an expansion in the training dataset size can lead to further improvements in model performance. For traditional text-based OEI, Bert-BiLSTM-CRF shows a 3.49\% increase in the F1 score after fine-tuning on the larger CIM-OEI corpus compared to training on CI-OEI. Similarly, Vicuna exhibits a performance boost of 1.33\% when fine-tuned on CIM-OEI versus CI-OEI. Nonetheless, the modest enhancement in the F1 score, despite significant increases in training data volume, indicates that reliance solely on textual input for solving the OEI task has inherent limitations.



Direct fine-tuning of STOEI-Vicuna on CIM-MOEI yields the best performance, surpassing Whispering-LLaMA by 13.29\%. This demonstrates the superior capability of STOEI to synergize speech and text features for Opinion Expression Identification compared to other multimodal approaches. Furthermore, a notable improvement of 5.36\% is observed compared to the results on the CI-MOEI dataset, indicating that synthesized speech also harbors a wealth of speech features beneficial for the OEI task. Regarding the suboptimal performance of models fine-tuned on text as a basis for STOEI, it is believed that text-based fine-tuning may lead to an over-reliance on the textual modality by the model.



\textbf{Fine-grain analysis of joint speech-text.}
In the analysis presented in Fig. \ref{fig:label_distribution}, the task becomes incrementally challenging for text-only-based LLMs. Specifically, in scenarios of brief sentences with limited contextual breadth, LLMs face difficulties in accurately identifying opinions and discerning their emotional polarities. However, with adequately lengthened contexts, LLMs exhibit improved performance, accurately interpreting the emotional content provided there is sufficient textual information. Nonetheless, when the sentence is too long, text-only-based is not enough for LLMs to perform more accurate OEI. This is because longer sentences contain more complex emotions and semantics, making it more difficult for LLMs to judge. Conversely, the data indicate that both multimodal combination methods yield higher accuracy rates for shorter sentences. This improvement is attributed to the enhanced clarity and prominence of tonal and emotional cues within shorter textual and speech inputs. As sentence length increases, the STOEI model demonstrates superior capability in discerning complex emotional expressions.

\usepgfplotslibrary{groupplots}
\usetikzlibrary{patterns,backgrounds}

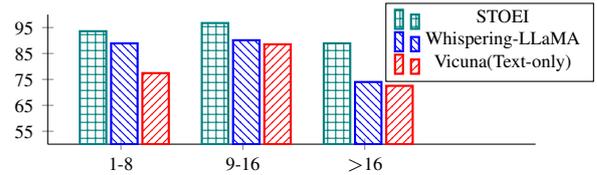
\begin{figure}[t]
\centering
\begin{tikzpicture}
    \pgfkeys{/pgf/number format/.cd,fixed, fixed zerofill,precision=2}
	\begin{groupplot}[group style={group name=myplot,group size=1 by 1,horizontal sep=15pt,vertical sep=30pt,xlabels at=edge bottom, ylabels at=edge left},height=4cm,width=8cm]

    \nextgroupplot[
    ybar,
    bar width=10pt,
    xtick={1.1,2.1,3.1},
     xticklabels={1-8,9-16,$>$16},
    xmax=4.5,xmin=0.5,
    x tick label style = {yshift=-0.5em, text height=0ex,font=\scriptsize},
    ytick = {55,65,75,85,95},
    yticklabels={55,65,75,85,95},
    ymax=100,ymin=50,
    y tick label style = {yshift=-0.3em, text height=0ex,font=\scriptsize},
	x label style = {font=\scriptsize},
	y label style = {font=\scriptsize},
    axis x line = bottom,
    axis y line=left,
    width = 8cm,
    height = 3.3cm,
    axis line style={-},
    title style={xshift=0em,yshift=-11em,font=\scriptsize},
    legend style={xshift=-1.5em, yshift=0.6em, text height=0ex, anchor=north, legend columns=1, font=\scriptsize},
    ]

    \addplot[teal, pattern=grid, thick, pattern color=teal, bar shift=-0.65em, bar width = 1.1em] coordinates
{
    (1,  93.59)
    (2,  96.74)
    (3,  88.92)

};\addlegendentry{STOEI}
\addplot[blue, pattern= north west lines, pattern color=blue, thick, bar shift=0.65em, bar width = 1.1em] coordinates
{
    (1,  88.89)
    (2,  90.11)
    (3,  74.00)
};\addlegendentry{Whispering-LLaMA}

\addplot[red, pattern= north east lines, pattern color=red, thick, bar shift=1.95em, bar width = 1.1em] coordinates
{
    (1,  77.42)
    (2,  88.52)
    (3,  72.52)
};\addlegendentry{Vicuna(Text-only)}

\end{groupplot}

\end{tikzpicture}
\caption{$\text{F}_1$ scores of Whispering-LLAMA, Vicuna(Text-only) and STOEI for different sentence lengths. The methods in the figure were all obtained by training directly on CIM-MOEI, except Vicuna(Text-only), which trained on CIM-OEI.}\label{fig:label_distribution}
\end{figure}

\textbf{Replacement large language model.} As can be seen from the table, the overall trend of the data is similar to that at Vicuna, but the results are not as good as Vicuna. We believe this disparity stems from the higher emotional intelligence of Vicuna, a viewpoint we have corroborated with the research of Wang et al. \cite{wang2023emotional}. And, the trend of the experiment is similar to Vicuna, validating the effectiveness of STOEI.

\begin{table}[ht]

  \centering
\caption{The experiment result of STOEI replacement the LLM. $\dagger$ indicates that the LLM has been fine-tuned on CIM-OEI.}
\label{tab:ala}
\small
  \setlength{\tabcolsep}{5pt}
\renewcommand{\arraystretch}{1.5} 
\resizebox{0.8\columnwidth}{!}{
  \begin{tabular}{c c c c c}
    \toprule
    
    \textbf{Base Model}& \textbf{Dataset} & \textbf{P} & \textbf{R} & \textbf{F1} \\
    \hline
 \multicolumn{5}{c}{\textbf{Traditional Text-based OEI}}\\
 \hline
 LLaMA2& CI-OEI& 67.20& 63.24&65.16\\
 LLaMA2& CIM-OEI& 71.28& 68.38&69.80\\
 \hline
 \multicolumn{5}{c}{\textbf{Text-Speech based MOEI}}\\
 \hline
    STOEI-LLaMA2& CI-MOEI& 0.7209 & 71.6& 71.83\\
    STOEI-LLaMA2$\dagger$& CIM-MOEI& 75.47& 72.88& 74.15\\
    STOEI-LLaMA2& CIM-MOEI& \textbf{83.39}& \textbf{80.17}& \textbf{81.75}\\
    \bottomrule
  \end{tabular}
}

\end{table}

\section{Conclusion}

In this work, we established the novel task of multimodal Opinion Expression Identification (MOEI), integrating text and speech to reflect real-world communication nuances.
By leveraging open-source datasets CMU MOSEI and IEMOCAP, we created the CI-MOEI dataset, addressing the alignment challenges between speech and text.
Additionally, we utilized Text-to-Speech technology on the MPQA dataset to form CIM-OEI, assessing the effectiveness of multimodal data as training material.
Employing large language models (LLMs), we developed a novel LLM-driven approach that combines speech and text modalities to help identify opinion expressions.
Our experiments demonstrated significant improvements in MOEI performance with this integrated approach, surpassing existing techniques by 9.20\% and achieving state-of-the-art (SOTA) results.
This study not only highlights the importance of combining textual and auditory information in sentiment analysis but also sets a precedent for future research in leveraging multimodal inputs for deeper emotional and opinion understanding.

Our work has two major limitations.
First, we only conducted our experiments on English datasets.
While our approach is applicable to other languages, there is a lack of experimental validation in these cases due to the time-consuming and high cost of data annotation.
Second, we have only compared a limited number of methods, and do not compare more methods for LLMs because of the large resource consumption.

\section{Acknowledgements}

This work is supported by the Project of Higher Education Stability Support Program (No.20220618160306001)





\bibliographystyle{IEEEtran}
\bibliography{mybib}

\end{document}